\documentclass[preprint,12pt]{elsarticle}

\usepackage{amsmath}
\usepackage{amssymb}
\usepackage{amsthm}
\usepackage{bm}
\usepackage{graphicx}
\usepackage{subcaption}
\usepackage{booktabs}
\usepackage{algorithm}
\usepackage{algpseudocode}
\usepackage{url}
\usepackage{cleveref}
\usepackage{lineno}
\theoremstyle{remark}
\newtheorem{remark}{Remark}
\journal{Journal of Computational Science}

\makeatletter
\newcommand{\safeincludegraphics}[2][]{%
  \begingroup
  \edef\@imgfile{\detokenize{#2}}%
  \IfFileExists{\@imgfile}{\includegraphics[#1]{\@imgfile}}{%
    \fbox{\parbox{0.9\linewidth}{Missing figure: \@imgfile}}%
  }%
  \endgroup
}
\makeatother

\newif\ifblindsubmission
\blindsubmissionfalse 

\begin{document}

\begin{frontmatter}

\title{Multirate Stein Variational Gradient Descent for Efficient Bayesian Sampling}

\ifblindsubmission
\author{Anonymous Author(s)}
\address{Anonymous Institution}
\else
\author[inst1]{Arash Sarshar\corref{cor1}}
\ead{arash.sarshar@csulb.edu}
\cortext[cor1]{Corresponding author}
\address[inst1]{Department of Computer Engineering and Computer Science, California State University, Long Beach, 90840, CA, USA}
\fi

\begin{abstract}
Many particle-based Bayesian inference methods use a single global step size
for all parts of the update. In Stein variational gradient descent (SVGD),
however, each update combines two qualitatively different effects: attraction
toward high-posterior regions and repulsion that preserves particle diversity.
These effects can evolve at different rates, especially in high-dimensional,
anisotropic, or hierarchical posteriors, so one step size can be unstable in
some regions and inefficient in others. We derive a multirate
version of SVGD that updates these components on different time scales. The
framework yields practical algorithms, including a symmetric split method, a
fixed multirate method (MR-SVGD), and an adaptive multirate method
(Adapt-MR-SVGD) with local error control. We evaluate the methods in a broad
and rigorous benchmark suite covering six problem families: a 50D Gaussian
target, multiple 2D synthetic targets, UCI Bayesian logistic regression,
multimodal Gaussian mixtures, Bayesian neural networks, and large-scale
hierarchical logistic regression. Evaluation includes posterior-matching metrics,
predictive performance, calibration quality, mixing, and explicit computational 
cost accounting.
Across these six benchmark families, multirate SVGD variants
improve robustness and quality-cost tradeoffs relative to vanilla SVGD. The
strongest gains appear on stiff hierarchical, strongly anisotropic, and
multimodal targets, where adaptive multirate SVGD is usually the strongest
variant and fixed multirate SVGD provides a simpler robust alternative at
lower cost.

\end{abstract}


\end{frontmatter}

\section{Introduction}

\label{sec:introduction}

Modern Bayesian inference often requires efficient sampling from complex, high-dimensional, and multimodal posterior distributions. As these problems grow in scale and complexity, traditional sampling methods face significant computational and statistical challenges. The advent of automatic differentiation has enabled the widespread use of gradient-based samplers, making methods such as stochastic gradient Langevin dynamics (SGLD) and stochastic gradient Hamiltonian Monte Carlo (SGHMC) practical and popular choices for large-scale applications \cite{welling2011sgld,chen2014sghmc}. While these approaches advance a single Markov chain using noisy gradients, an alternative paradigm is offered by interacting particle methods, which evolve a population of particles whose empirical distribution approximates the target posterior. This population-based view is central to particle-based variational inference \cite{pmlr-v97-liu19i}. Within this class, SVGD is a prominent deterministic representative, and recent work has examined acceleration, particle degeneracy, and kernel design in related Stein-particle methods \cite{pmlr-v97-liu19i,pmlr-v80-zhuo18a,wang2019matrixsvgd}. It opens new avenues for improving sampling efficiency and robustness, especially in challenging inference settings.

A useful unifying perspective is to view interacting particle methods as
discretizations of a particle flow (possibly with a stochastic component) that
transports an initial ensemble toward the target distribution \cite{Sandu_2025_EnsembleFokkerPlanck}.
Here, we focus on Stein variational gradient descent (SVGD) \cite{liu2016svgd} as a
representative interacting particle method. SVGD constructs a velocity field
for the particles based on the target density and a reproducing kernel Hilbert
space (RKHS) structure, and updates the particles by discretizing this flow \cite{liu2016svgd}.
The velocity field naturally separates into an attractive drift, which moves
particles toward high-density regions, and a repulsive interaction, which
maintains particle diversity and support coverage of multiple modes.

This drift-repulsion structure creates the main numerical challenge studied in
this paper. When particles crowd, repulsion can become stiff and destabilize
updates. On ther hand, the drift can require finer resolution on anisotropic targets to
follow the geometry of $\log p$ closely. A single step size must therefore
balance competing requirements and can either over-damp repulsion, leading to
loss of diversity and mode coverage, or under-resolve drift, leading to slow
progress.

Moreover, repulsion typically requires kernel evaluation at each
step, so increasing the stability by reducing  the step-size can become
computationally expensive. Finally, repulsive forces can weaken in high
dimensions or under limited particle resolution, leading to particle collapse
and poor support of multiple modes \cite{pmlr-v80-zhuo18a,pmlr-v108-zhang20d}.
In practice, the resulting single-step-size compromise is both statistically
and computationally inefficient: it can degrade support coverage and predictive
quality while still spending substantial kernel work to stabilize the wrong
part of the update.


A common approach to solving systems with significantly different 
components is to use partitioned methods such as implicit-explicit (IMEX)  \cite{Pareschi_2000,sarshar2020parimex} and
multirate integration methods \cite{Sandu_2016_MR-GARK}.  In the context of ordinary
differential equations (ODEs), multirate methods exploit a natural splitting of the dynamics into 
fast and slow components, allowing each to be integrated with a step size appropriate
to its scale \cite{sarshar2019mrgark,sarshar2021implicitgark,sarshar2021glmadi}. 
For example, in a split ODE of the form $\dot{x} = \phi_{\text{fast}}(x) + \phi_{\text{slow}}(x)$, multirate schemes
apply more frequent, smaller steps to the stiff or rapidly varying part, while updating the
slow part less often. This approach improves both stability and efficiency in multi-scale
dynamical systems, and has been successfully applied in a variety of scientific computing contexts \cite{sarshar2020parimex,sarshar2022surrogate}. 

Motivated by advances in multirate time-integration
for multi-scale dynamical systems—including MrGARK, IMEX, and related splitting schemes \cite{sarshar2019mrgark,sarshar2021implicitgark,sarshar2021glmadi} we propose to integrate the drift and repulsion terms on separate time scales. This leads to a number of multirate SVGD variants, including symmetric splitting, fixed repulsion substepping, and adaptive error-controlled drift substepping.

To assess the effectiveness of these methods, we conduct a comprehensive empirical evaluation. We compare our multirate SVGD variants with established approaches from the literature across a range of benchmark problems, using a variety of task-appropriate metrics. Results highlight the strengths and limitations of each method under diverse geometric and statistical challenges.

\subsection*{Contributions}

The main contributions of this work are:
\begin{itemize}
  \item Multirate SVGD formulations that explicitly separate drift and repulsion
  and support symmetric splitting and fixed multirate schedules.
  \item An adaptive error-controlled multirate SVGD variant that selects 
  substeps based on a local error estimator while keeping repulsion stable.
  \item A broad empirical study showing that multirate SVGD improves robustness
  and quality-cost tradeoffs compared to SVGD and representative chain methods, with the strongest gains on
  anisotropic, multimodal, and hierarchical targets.
  \item A benchmark suite spanning many different targets,
  equipped with appropriate diagnostics, early stopping, performance evaluation metrics for efficiency comparisons.
\end{itemize}

We present our work as follows. Section~\ref{sec:method} presents the SVGD
variants and the multirate construction. Section~\ref{sec:exp_setup} describes
the benchmarks, diagnostics, and experimental protocol. Section~\ref{sec:results}
reports the results. Section~\ref{sec:conclusions} summarizes the findings and
discusses directions for future work.

\section{Method}
\label{sec:method}

\subsection{Background: SVGD as a particle flow}

Let $p(x)$ denote a target density on $\mathbb{R}^d$ and let
$\{x_i^t\}_{i=1}^N$ be a set of particles at iteration $t$.  SVGD evolves particles by following a deterministic
interacting-particle flow whose velocity field is constrained to RKHS \cite{liu2016svgd}. A first-order discretization of
this flow is the update
\begin{equation}
  x_i^{t+1} = x_i^t + h\,\phi(x_i^t),
  \label{eq:svgd-update}
\end{equation}
where $h>0$ is a step size and the SVGD velocity field is
\begin{equation}
  \phi(x) = \frac{1}{N}\sum_{j=1}^N
  \Big[ k(x_j, x)\,\nabla_{x_j}\log p(x_j) + \nabla_{x_j} k(x_j, x) \Big].
  \label{eq:svgd-velocity}
\end{equation}
The first term in \cref{eq:svgd-velocity} is an \emph{attractive} drift that
pushes particles toward high-density regions via the score $\nabla \log p$. The second term is a \emph{repulsive} interaction that discourages particle collapse.
The multirate constructions below are kernel-agnostic: they apply once a
kernel-induced attractive field and repulsive field have been specified.

\subsection{Drift--repulsion splitting and multirate integration}

The central numerical observation motivating this work is that the drift and
repulsion components of \cref{eq:svgd-velocity} can impose different stability
constraints and can naturally evolve on different time scales. We therefore
write the SVGD flow as a split system
\begin{equation}
  \dot{x} = f_{\text{rep}}(x) + f_{\text{drift}}(x),
  \label{eq:svgd-splitting}
\end{equation}
where, for particle $i$,
\begin{align}
  f_{\text{drift}}(x_i)
  &:= \frac{1}{N}\sum_{j=1}^N k(x_j, x_i)\,\nabla_{x_j}\log p(x_j),
  \label{eq:svgd-drift-term}\\
  f_{\text{rep}}(x_i)
  &:= \frac{1}{N}\sum_{j=1}^N \nabla_{x_j} k(x_j, x_i).
  \label{eq:svgd-rep-term}
\end{align}
We next define multirate discretizations that apply different substepping
strategies to $f_{\text{drift}}$ and $f_{\text{rep}}$ while keeping a common
macro-step size $h$.

\paragraph{Strang-split SVGD}

Define the forward Euler update that advance the split system by a step size
$\tau$:
\begin{equation*}
  \Psi_{\text{rep}}^{\tau}(x) = x + \tau f_{\text{rep}}(x),
  \qquad
  \Psi_{\text{drift}}^{\tau}(x) = x + \tau f_{\text{drift}}(x).
\end{equation*}
\emph{Strang
splitting} \cite{strang1968difference} is a classical second-order method that
constructs a macro-step of size $h$ by composing such
substeps. In our setting, one macro step takes the form
\begin{equation}
  x^{t+1}
  = \Psi_{\text{rep}}^{h/2}\circ\Psi_{\text{drift}}^{h}\circ\Psi_{\text{rep}}^{h/2}(x^{t}),
  \label{eq:strang-split-svgd}
\end{equation}
which alternates half a repulsion step, a full drift step, and another half
repulsion step. In practice, the half-steps can be implemented as
$\frac{h}{2m}$ smaller substeps for stability.

\paragraph{Fixed multirate SVGD (MR-SVGD)}
Fixed multirate integrators provide a principled way to allocate different step
sizes to fast and slow components of a split system
\cite{Sandu_2016_MR-GARK,sarshar2019mrgark,sarshar2021implicitgark}. In our SVGD
setting, however, higher-order multirate schemes are often not cost-effective:
their additional stages translate into multiple evaluations of the drift and
repulsion fields per macro step, and each such evaluation is expensive due to
gradient computations and, in particular, $O(N^2)$ kernel interactions.
Moreover, practical accuracy is also limited by finite-particle effects and by
choices such as bandwidth selection and early stopping, which can mute the
benefits of increasing time-integration order at a fixed computational budget.
We therefore utilize a first-order multirate construction using Euler updates that preserves a macro step of size $h$ for the drift while resolving the
repulsion with $m$ substeps:
\begin{equation}
  x^{t+1}
  = \Psi_{\text{drift}}^{h}\circ\Big(\Psi_{\text{rep}}^{h/m}\Big)^{m}(x^{t}).
  \label{eq:mr-svgd}
\end{equation}
This retains a coarse drift step while stabilizing the repulsive interaction
when particles crowd, at the cost of additional kernel evaluations. The
corresponding implementation-level procedure is summarized in
\cref{alg:mr_svgd}.

\begin{algorithm}[t]
\caption{Fixed multirate SVGD (MR-SVGD)}
\label{alg:mr_svgd}
\begin{algorithmic}[1]
\Require Particles $\{x_i\}_{i=1}^N$, step size $h$, repulsion substeps $m$, kernel $k$
\State $x \leftarrow \{x_i\}$
\For{$s=1$ to $m$}
  \State compute repulsion $f_{\text{rep}}(x)$ using \cref{eq:svgd-rep-term}
  \State $x \leftarrow x + (h/m)\,f_{\text{rep}}(x)$
\EndFor
\State compute drift $f_{\text{drift}}(x)$ using \cref{eq:svgd-drift-term}
\State $x \leftarrow x + h\,f_{\text{drift}}(x)$
\State \Return $x$
\end{algorithmic}
\end{algorithm}

\paragraph{Adaptive multirate SVGD (Adapt-MR-SVGD)}
Although fixed-$m$ schedules are simple and robust, they impose a uniform drift
resolution along the entire trajectory. As a result, they may spend unnecessary
computational effort in benign regions while providing insufficient resolution
in locally stiff regimes. To address this imbalance, we choose the number of
drift microsteps $m$ adaptively at each macro step, while retaining one
repulsion update per macro step. This follows the standard local-error-control
perspective used in ODE
integrators, where local error estimates are used to adjust resolution
\cite[Ch. II]{Hairer_book_I}. After the repulsion update, denote the state by
$x=\Psi_{\text{rep}}^{h}(x^{t})$. We estimate the local
drift-discretization error by comparing one drift step of size
$h$ with two consecutive drift half-steps of size $h/2$,
\begin{align}
  x_{\text{full}} &= x + h\,f_{\text{drift}}(x), \\
  x_{\text{half}} &= x + \tfrac{h}{2}\,f_{\text{drift}}(x), \\
  \tilde{x}_{\text{full}}  &= x_{\text{half}} + \tfrac{h}{2}\,f_{\text{drift}}(x_{\text{half}}).
\end{align}
We then form the relative discrepancy
\begin{equation}
  \rho
  = \frac{\|\tilde{x}_{\text{full}} - x_{\text{full}}\|_2}{\|x\|_2 + \epsilon},
  \label{eq:adapt-mr-error}
\end{equation}
which serves as a dimensionless local error indicator for the drift update. $\epsilon>0$ is a small constant to avoid division by zero.
For a first-order drift integrator, this local error behaves
as \cite[Ch. II.4]{Hairer_book_I}:
\[
\rho(h) = C h^2 + \mathcal{O}(h^3),
\]
where $C$ is the principal error constant dependent on $f$ and its derivatives. If the drift step is replaced by $m$ microsteps of size $h/m$, then
\[
\rho_m \approx C(h/m)^2 \approx \rho/m^2.
\]
Imposing the target local error tolerance $\rho_m \approx \text{Tol}$ gives
$m \approx \sqrt{\rho/\text{Tol}}$. We therefore use the clipped integer
controller
\begin{equation}
  m \leftarrow \mathrm{clip}\Big(\Big\lceil \sqrt{\rho/\text{Tol}}\Big\rceil,
  m_{\min}, m_{\max}\Big),
  \label{eq:adapt-mr-m}
\end{equation}
which increases $m$ when $\rho>\text{Tol}$ and decreases it when
$\rho\le\text{Tol}$, while enforcing the robustness bounds
$m_{\min}\le m\le m_{\max}$.
The macro-step update is then
\begin{equation}
  x^{t+1}
  = \Big(\Psi_{\text{drift}}^{h/m}\Big)^m\circ\Psi_{\text{rep}}^{h}(x^t),
  \qquad m = m(x^t,h).
  \label{eq:adapt-mr-step}
\end{equation}
This
strategy increases drift resolution only when the local error indicator suggests
stiffness or rapid variation in the drift component.  \Cref{alg:adapt_mr_svgd} provides the complete algorithm for this method.

\begin{remark}
In practice, if $m\le 2$ we can reuse the already computed, locally-higher-order solution $\tilde{x}_{\text{full}} $ to avoid extra drift evaluations.
\end{remark}

\begin{algorithm}[t]
\caption{Adaptive multirate SVGD (Adapt-MR-SVGD)}
\label{alg:adapt_mr_svgd}
\begin{algorithmic}[1]
\Require Particles $\{x_i\}_{i=1}^N$, step size $h$, bounds $m_{\min}, m_{\max}$, tolerance $\text{Tol}$
\State $x \leftarrow \{x_i\}$
\State compute repulsion $f_{\text{rep}}(x)$ using \cref{eq:svgd-rep-term}
\State $x \leftarrow x + h\,f_{\text{rep}}(x)$
\State compute drift $f_{\text{drift}}(x)$ using \cref{eq:svgd-drift-term}
\State $x_{\text{full}} \leftarrow x + h\,f_{\text{drift}}(x)$
\State $x_{\text{half}} \leftarrow x + (h/2)\,f_{\text{drift}}(x)$
\State compute drift $f_{\text{drift}}(x_{\text{half}})$
\State $\tilde{x}_{\text{full}} \leftarrow x_{\text{half}} + (h/2)\,f_{\text{drift}}(x_{\text{half}})$
\State estimate relative error $\rho = \|\tilde{x}_{\text{full}} - x_{\text{full}}\| / (\|x\| + \epsilon)$
\State $m \leftarrow \mathrm{clip}(\lceil \sqrt{\rho/\text{Tol}} \rceil, m_{\min}, m_{\max})$
\If{$m \le 2$}
  \State $x \leftarrow \tilde{x}_{\text{full}}$ \Comment{reuse current predictor if accurate}
\Else
  \For{$s=1$ to $m$}
    \State compute drift $f_{\text{drift}}(x)$
    \State $x \leftarrow x + (h/m)\,f_{\text{drift}}(x)$
  \EndFor
\EndIf
\State \Return $x$
\end{algorithmic}
\end{algorithm}

\section{Experimental Setup}
\label{sec:exp_setup}

We evaluate multirate SVGD variants on six benchmark groups: a 50D Gaussian
target, a suite of 2D synthetic targets, UCI logistic regression, a 2D Gaussian
mixture (mix8), a one-hidden-layer Bayesian neural network (BNN), and a
large-scale hierarchical logistic-regression (HLR) benchmark. The comparison
focuses on both statistical quality and computational cost. This section defines
the shared diagnostics and protocol.

\subsection{Evaluation Metrics}

Evaluation combines distributional fidelity, predictive performance, mixing, and
explicit computational cost accounting. Metrics used across multiple
benchmarks are defined here once. Benchmark-specific metrics are introduced
in their respective sections.

\paragraph{Moment fidelity}
When reference moments are available, we report
$e_\mu=\|\hat\mu-\mu_{\mathrm{ref}}\|_2$ and
$e_\Sigma=\|\hat\Sigma-\Sigma_{\mathrm{ref}}\|_F$, where
$\hat\mu=\frac{1}{N}\sum_{i=1}^N x_i$ and
$\hat\Sigma=\frac{1}{N}\sum_{i=1}^N(x_i-\hat\mu)(x_i-\hat\mu)^\top$.
These summarize first- and second-order bias, which is critical in the
Gaussian and synthetic-target settings.

\paragraph{Stein discrepancy}
For distributional fidelity, we use kernel Stein discrepancy (KSD) with score
$s_p(x)=\nabla_x \log p(x)$ and an RBF base kernel $k_h(\cdot,\cdot)$
\cite{liu2016ksd,pmlr-v70-gorham17a}. The corresponding Stein kernel is
\begin{equation}
\begin{aligned}
\kappa_p(x,x')={}&s_p(x)^\top s_p(x')\,k_h(x,x')
+s_p(x)^\top \nabla_{x'} k_h(x,x') \\
&+s_p(x')^\top \nabla_x k_h(x,x')
+\operatorname{tr}\!\left(\nabla_x\nabla_{x'}k_h(x,x')\right),
\end{aligned}
\label{eq:ksd-stein-kernel}
\end{equation}
and the empirical discrepancy is
\begin{equation}
\operatorname{KSD}(X;p)=
\left(\frac{1}{N^2}\sum_{i=1}^{N}\sum_{j=1}^{N}\kappa_p(x_i,x_j)\right)^{1/2}.
\label{eq:ksd-metric}
\end{equation}
KSD measures how far the empirical particle distribution is from satisfying the
target's Stein identity. Intuitively, it acts as a global goodness-of-fit
metric: it is small when the particles collectively match the target law, and
large when they exhibit systematic errors such as bias, incorrect spread, or
missed modes.

\paragraph{Mean log-density fit}
We complement KSD with the empirical mean log-density
\begin{equation}
\overline{\log p}(X)=\frac{1}{N}\sum_{i=1}^{N}\log p(x_i),
\label{eq:mean-logp-metric}
\end{equation}
which measures how strongly the particles concentrate in high-density regions
of the target. Unlike KSD, this is a pointwise criterion: it rewards samples
for lying in locally plausible regions, but by itself it does not certify
global distributional agreement. Reporting
\cref{eq:ksd-metric,eq:mean-logp-metric} therefore separates global
distributional fidelity from local density concentration.

\paragraph{Predictive performance and calibration}
For the binary-classification experiments (UCI and BNN tasks), we report
accuracy, negative log-likelihood, and
expected calibration error \footnote{Accuracy as the fraction of correctly classified examples
under a $0.5$ threshold on the posterior predictive probability.  $\mathrm{acc}(\mathcal{B}_b)$  is the empirical accuracy in bin $\mathcal{B}_b$ under the same threshold, and $\mathrm{conf}(\mathcal{B}_b)$ is
the mean predictive probability in that bin.},
\begin{align}
\mathrm{NLL}
&= -\frac{1}{M}\sum_{m=1}^M
\big[y_m\log \hat p_m+(1-y_m)\log(1-\hat p_m)\big], \\
\mathrm{ECE}
&= \sum_{b=1}^{B}\frac{|\mathcal{B}_b|}{M}\,
\big|\mathrm{acc}(\mathcal{B}_b)-\mathrm{conf}(\mathcal{B}_b)\big|.
\end{align}
In all predictive experiments, ECE is computed with $B=10$ equal-width bins on
$[0,1]$. Together, these metrics capture thresholded classification
performance, probabilistic predictive fit, and calibration quality.

\paragraph{Mixing efficiency}
To quantify dependence, we report ESS as a supplemental univariate mixing
diagnostic, computed from the first coordinate using a standard
autocorrelation-based estimator with initial-positive truncation
\cite{geyer1992practical}. For particle methods at a checkpoint, the diagnostic
sequence is formed from the first coordinate across particles; for single-chain
baselines, it is formed from a rolling window of recent iterates. Because this
quantity is one-dimensional, we do not interpret ESS as a full multivariate
efficiency measure but as a complementary statistic.

\paragraph{Cost accounting}
Alongside wall-clock time, we track gradient and kernel evaluation counts to
support hardware-agnostic efficiency comparisons across methods with different workloads.

\begin{remark}
For the common metrics above, lower is better for $e_\mu$, $e_\Sigma$, KSD,
NLL, ECE, wall-clock time, and gradient- and kernel-evaluation counts,
whereas higher is better for ESS, accuracy, and mean log-probability.
\end{remark}

\subsection{Experimental Protocol}

Across benchmarks, we compare the same method family whenever applicable:
vanilla SVGD, Strang-split SVGD, fixed multirate SVGD (MR-SVGD), adaptive
multirate SVGD (Adapt-MR-SVGD), and the single-chain baselines SGLD and an
SGHMC-style method. The SVGD-family updates are those defined in
\cref{sec:method}, namely
\cref{eq:svgd-update,eq:strang-split-svgd,eq:mr-svgd,eq:adapt-mr-step}.

Single-chain baselines use the standard stochastic-gradient update
\begin{equation}
x_{t+1}=x_t+\eta\nabla \log p(x_t)+\sqrt{2\eta}\,\xi_t,\qquad \xi_t\sim\mathcal{N}(0,I),
\label{eq:sgld-update}
\end{equation}
for SGLD \cite{welling2011sgld}, and
\begin{equation}
v_{t+1}=\beta v_t+\eta\nabla \log p(x_t)+\sqrt{2\alpha\eta}\,\xi_t,\qquad x_{t+1}=x_t+v_{t+1},
\label{eq:sghmc-update}
\end{equation}
for an SGHMC-style baseline \cite{chen2014sghmc}. Here $\eta$ is the step
size, $\alpha$ is a friction or noise parameter, and $\beta\in(0,1)$ is the
momentum decay.

\paragraph{Particle configuration and kernel choice}
For particle methods, we use $N=128$ particles; for single-chain methods, we
compute metrics from a rolling window of recent iterates to obtain comparable
finite-sample summaries. Unless stated otherwise, particle methods use a
Gaussian RBF kernel with bandwidth selected by the median heuristic and a
benchmark-specific scaling factor. For Mix8, we instead use a shared
multiscale RBF kernel with bandwidth factors $\{0.5,1,2\}$ to better probe
local spacing and longer-range interactions across modes.

\paragraph{Implementation safeguards}
To improve numerical robustness, the multirate implementations clip raw score
components to the interval $[-50,50]$ before assembling the update. Benchmarks
with pronounced instability additionally use a short non-finite fail-fast
guard, which we note locally when it is active.

\paragraph{Checkpointing and model selection}
Unless noted otherwise, experiments run for up to 1,000 iterations over five
seeds with checkpointed evaluation every 20 iterations. All benchmarks use a
patience-based early-stopping rule triggered by persistent degradation of a
designated monitor metric. Pure sampling benchmarks use KSD as the monitor
(50D Gaussian, 2D synthetic targets, and Mixture2D), whereas predictive
benchmarks use held-out NLL (UCI logistic regression, BNN, and HLR). For each
run, we retain the best finite checkpoint under that monitor before stopping.
Accuracy and ECE are reported for predictive tasks but are never used as
stopping criteria.

\section{Results}
\label{sec:results}
\label{sec:experiments}

In the following, we introduce each benchmark, explain its relevance to
particle-based sampling, and report the corresponding empirical results. The
benchmark suite consists of a 50D Gaussian target, 2D synthetic targets, a
multimodal Gaussian mixture (Mix8), UCI logistic regression, a Bayesian neural
network, and large-scale hierarchical logistic regression.

\subsection{50D Gaussian}

This benchmark tests the stability of sampling in a high-dimensional setting with anisotropic
covariance structure. We run 1,000 sampling iterations with checkpointing every 50 steps
and a 5-seed ablation. Reported metrics are moment errors
($\|\hat{\mu}-\mu\|_2$ and $\|\hat{\Sigma}-\Sigma\|_F$), ESS, KSD, and time-to best-checkpoint.
\begin{figure}[t]
  \centering
  \safeincludegraphics[width=0.9\linewidth]{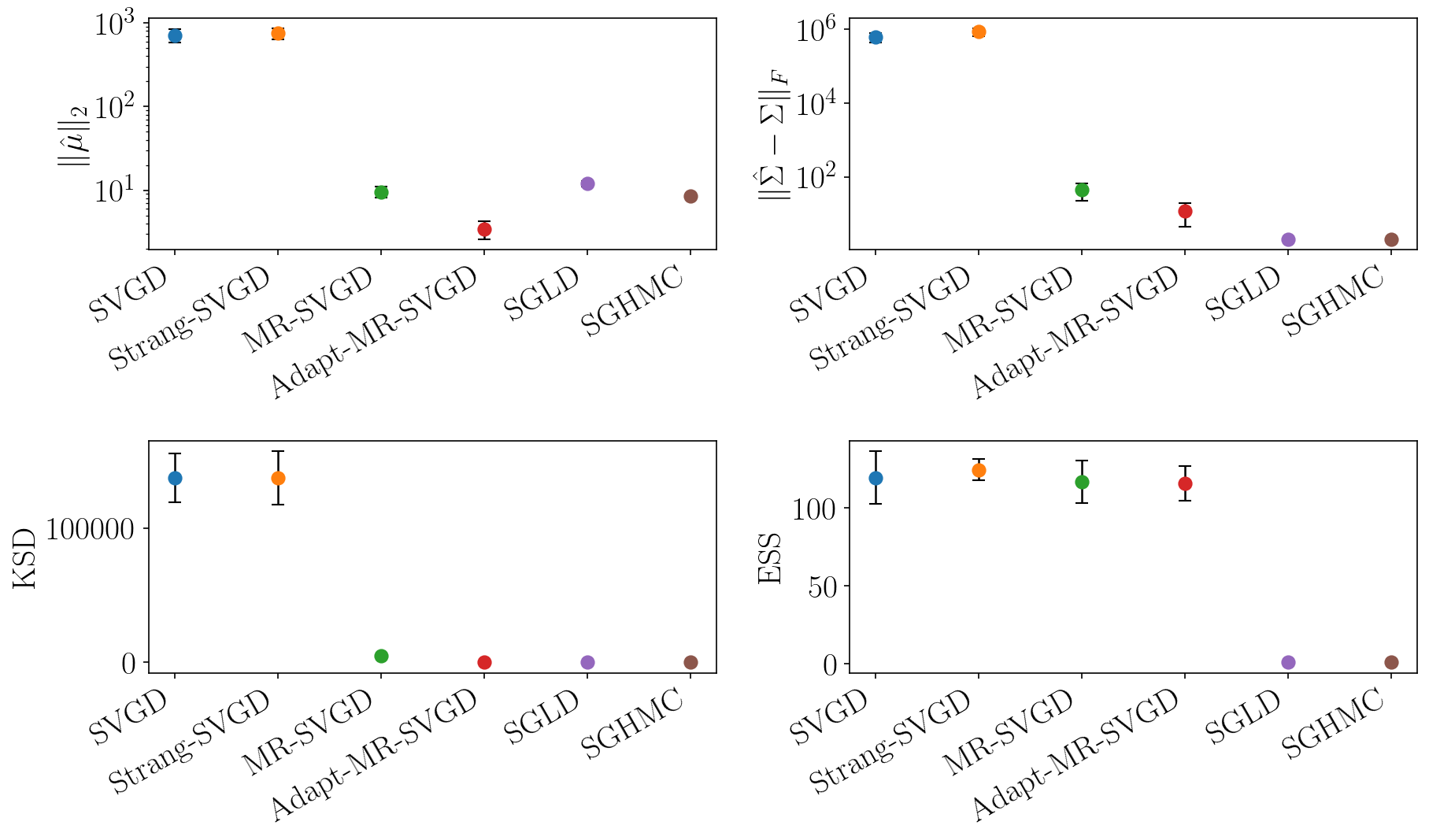}
  \caption{50D Gaussian final-checkpoint summary across methods. The four panels
  report $\|\hat{\mu}-\mu\|_2$, $\|\hat{\Sigma}-\Sigma\|_F$, KSD, and ESS,
  respectively. Markers show mean values across seeds and error bars indicate
  one standard deviation. Lower is better for the first three metrics, whereas
  higher is better for ESS. This figure highlights robustness differences under
  strong anisotropy: Adapt-MR-SVGD is the only particle variant that keeps both
  moment errors and KSD under control.}
  \label{fig:gauss50_summary}
\end{figure}

\begin{figure}[t]
  \centering
  \safeincludegraphics[width=0.9\linewidth]{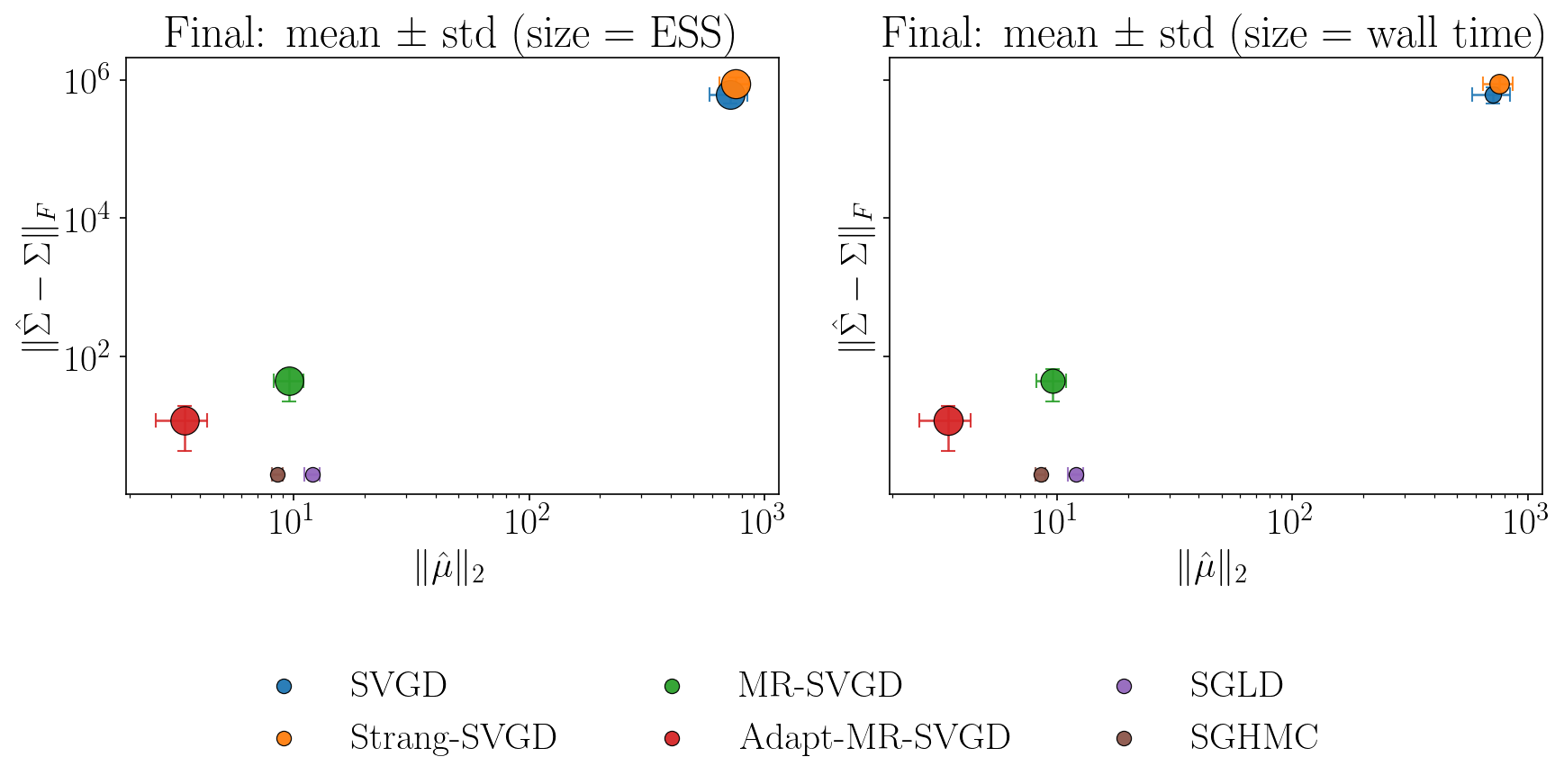}
  \caption{50D Gaussian: final mean $\pm$ std Pareto plot in moment-error space
  with marker size encoding ESS (left) and wall time (right).}
  \label{fig:gauss50_pareto}
\end{figure}

Among the methods, \cref{fig:gauss50_summary} shows that
Adapt-MR-SVGD is the only variant that keeps both moment errors and KSD under
control in this anisotropic Gaussian setting: it substantially improves on
MR-SVGD and avoids the severe degradation exhibited by vanilla SVGD and
Strang-SVGD. The chain baselines remain competitive on some final-checkpoint
fidelity metrics, but they do so with ESS near $4$, far below the particle
methods. Within the particle family, \cref{fig:gauss50_pareto} shows that MR-SVGD offers
a cheaper but less accurate baseline, whereas Adapt-MR-SVGD
achieves the strongest quality at a higher computational cost.

\subsection{2D synthetic targets}
\begin{figure}[t]
  \centering
  \begin{subfigure}[t]{0.84\linewidth}
    \centering
    \safeincludegraphics[width=\linewidth]{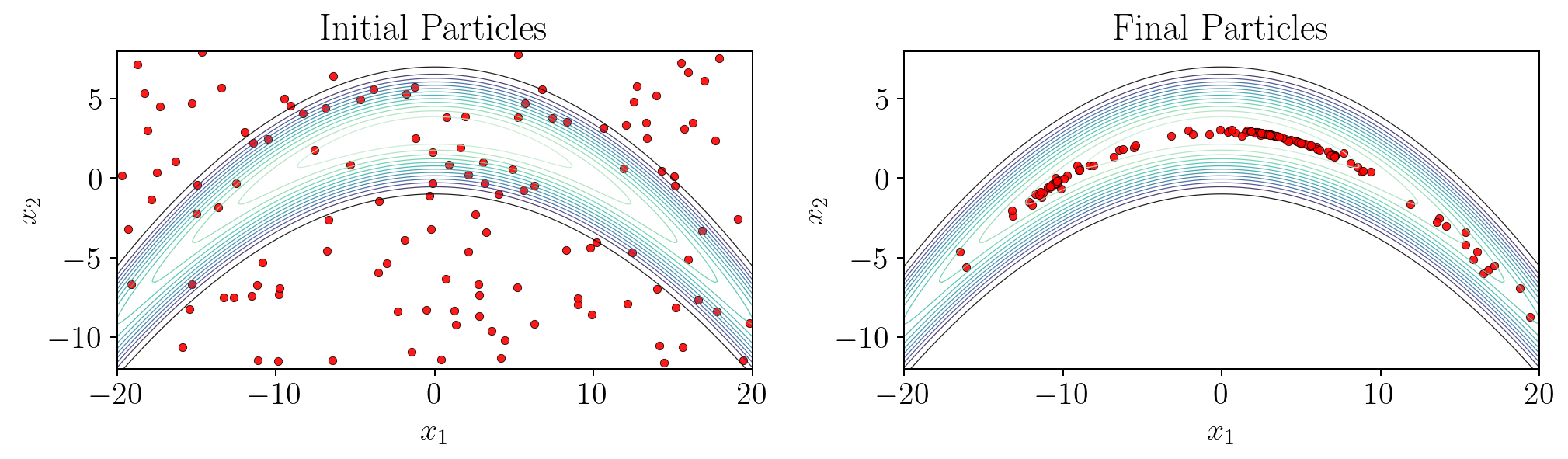}
    \caption{Banana.}
    \label{fig:2d_demo_banana}
  \end{subfigure}

  \vspace{0.15em}

  \begin{subfigure}[t]{0.84\linewidth}
    \centering
    \safeincludegraphics[width=\linewidth]{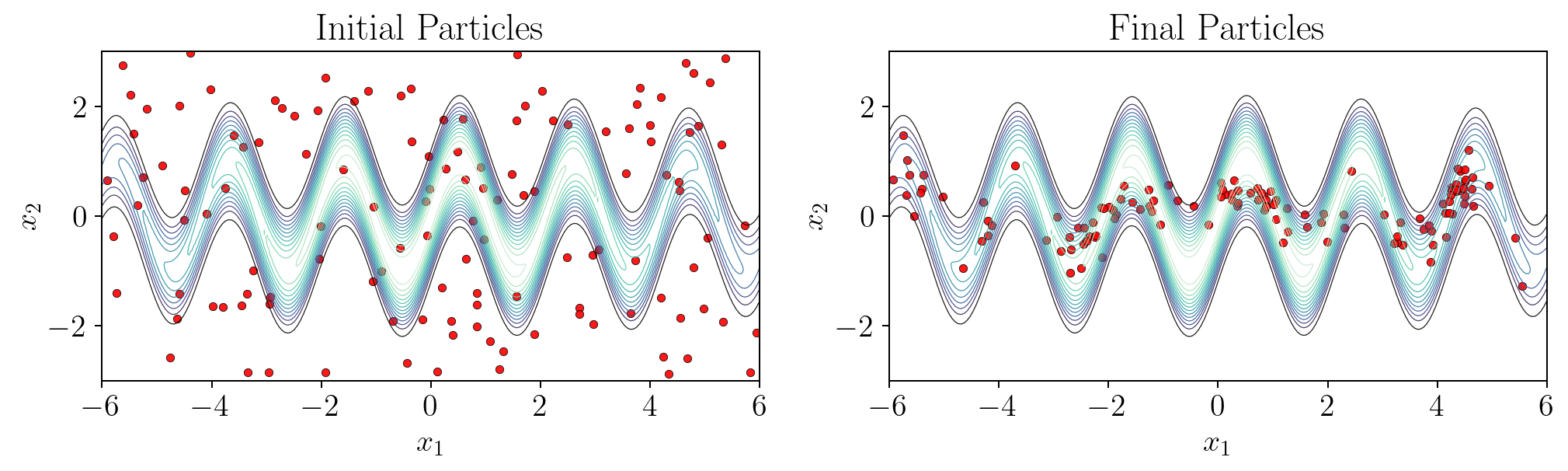}
    \caption{Squiggly.}
    \label{fig:2d_demo_squiggly}
  \end{subfigure}

  \vspace{0.15em}

  \begin{subfigure}[t]{0.84\linewidth}
    \centering
    \safeincludegraphics[width=\linewidth]{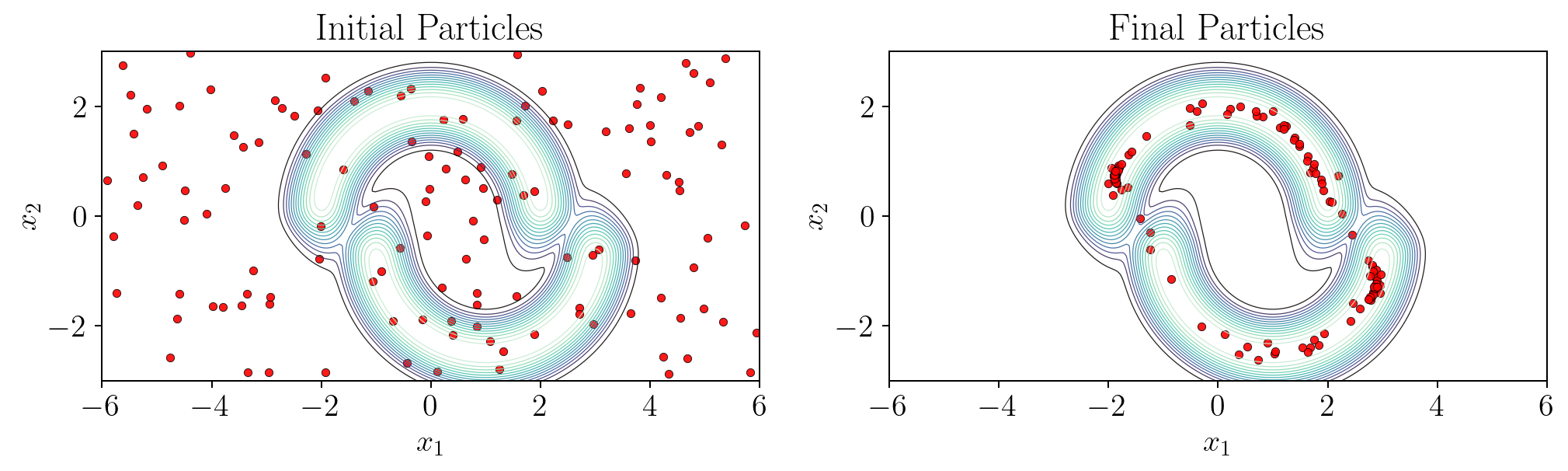}
    \caption{Two moons.}
    \label{fig:2d_demo_two_moons}
  \end{subfigure}

  \vspace{0.15em}

  \begin{subfigure}[t]{0.84\linewidth}
    \centering
    \safeincludegraphics[width=\linewidth]{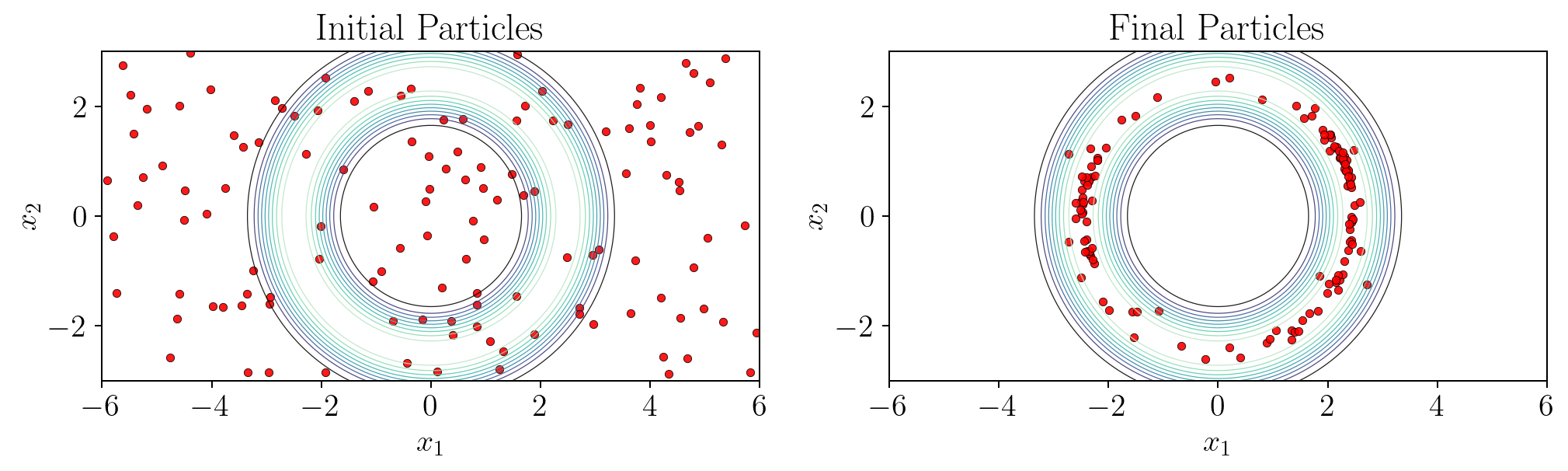}
    \caption{Ring.}
    \label{fig:2d_demo_ring}
  \end{subfigure}

  \caption{2D target visualization panels produced with Adapt-MR-SVGD under
  visualization-only settings. Each subfigure shows target-density contours with
  initial particles (left) and short-run final particles (right).}
  \label{fig:2d_demo_panel}
\end{figure}

In this benchmark, we evaluate a suite of 2D targets with varying geometry and multimodality, including banana, ring, squiggly, two-moons, and funnel.
To contextualize the geometry of these targets, \cref{fig:2d_demo_panel} shows
short-run visualization panels for representative cases.
Here KSD serves both as the stop monitor and the
primary quality metric. We add a safety guard against non-finite KSD values in
the stopping policy, because several targets can produce unstable particle
trajectories. The main comparison in \cref{tab:2d_summary} focuses on KSD, mean
log probability, gradient-evaluation cost, and wall time. The panels in
\cref{fig:2d_demo_panel} are qualitative only and are included to illustrate
target geometry and short-run particle behavior.

\begin{table}[t]
  \centering
  \scriptsize
  \setlength{\tabcolsep}{4pt}
  \begin{tabular}{lcccc}
    \toprule
    Method & Best KSD$\downarrow$ & mean logp@best$\uparrow$ & grad@best$\downarrow$ & wall@best (s)$\downarrow$ \\
    \midrule
    \multicolumn{5}{l}{\textit{Banana}} \\
    Strang-SVGD & 0.114 & -0.426 & 100 & 1.35 \\
    MR-SVGD & 0.114 & -0.392 & 80 & 1.13 \\
    Adapt-MR-SVGD & 0.114 & -0.393 & 160 & 0.80 \\
    SVGD & 0.114 & -0.391 & 80 & 0.32 \\
    \addlinespace[0.3em]
    \multicolumn{5}{l}{\textit{Funnel}} \\
    MR-SVGD & 0.256 & -3.136 & 160 & 1.24 \\
    Adapt-MR-SVGD & 0.357 & -0.197 & 84 & 0.40 \\
    Strang-SVGD & 0.369 & -9.289 & 120 & 1.46 \\
    SVGD & 0.377 & -6.801 & 160 & 0.46 \\
    \addlinespace[0.3em]
    \multicolumn{5}{l}{\textit{Ring}} \\
    Adapt-MR-SVGD & 0.226 & -0.004 & 238 & 0.82 \\
    MR-SVGD & 64.35 & -815.6 & 50 & 0.70 \\
    SVGD & 131.6 & -3753 & 20 & 0.16 \\
    Strang-SVGD & 665.7 & $-7.92\times 10^5$ & 240 & 2.79 \\
    \addlinespace[0.3em]
    \multicolumn{5}{l}{\textit{Squiggly}} \\
    Adapt-MR-SVGD & 0.719 & -0.910 & 826 & 2.02 \\
    Strang-SVGD & 22.49 & $-5.25\times 10^4$ & 200 & 2.47 \\
    SVGD & 22.60 & $-2.81\times 10^4$ & 80 & 0.32 \\
    MR-SVGD & 54.04 & -422.7 & 100 & 1.36 \\
    \addlinespace[0.3em]
    \multicolumn{5}{l}{\textit{Two moons}} \\
    Adapt-MR-SVGD & 0.315 & -0.009 & 182 & 0.67 \\
    MR-SVGD & 134.3 & -1250 & 40 & 0.60 \\
    Strang-SVGD & 1892 & $-1.84\times 10^6$ & 120 & 1.55 \\
    SVGD & 2085 & $-2.58\times 10^6$ & 80 & 0.34 \\
    \bottomrule
  \end{tabular}
  \caption{2D particle-method summary over five seeds. For each target and
  method, we report medians over seeds at the best finite-KSD checkpoint.
  Methods are ordered by median KSD within each target block. Banana shows
  little separation, funnel remains nuanced, and Adapt-MR-SVGD is strongest on
  ring, squiggly, and two moons.}
  \label{tab:2d_summary}
\end{table}

\Cref{tab:2d_summary} shows a sumamry of the metrics for the 2D suite.
Banana is effectively a tie: all four particle methods reach nearly identical
best KSD values, with only minor differences in mean log probability and cost.
The other targets separate more sharply. On ring, squiggly, and two moons,
Adapt-MR-SVGD is the only method that stays near the target while keeping KSD
below $1$, whereas the other particle methods remain orders of magnitude worse
in both KSD and mean log probability. MR-SVGD is often cheaper in
gradient-evaluation count on those targets, but it pays for that reduction with
substantially weaker fidelity.

Funnel is the main exception. There MR-SVGD attains the best median KSD, but
Adapt-MR-SVGD has much better median mean log probability while also using
fewer gradient evaluations and less wall time. Taken together, these results
show that adaptive multirate splitting is not uniformly best on every benign 2D
target, but it is the only particle variant that combines strong
distributional fidelity with competitive cost across the harder geometries.

\subsection{Mixture2D (mix8)}

Mix8 is an equal-weight eight-component Gaussian mixture with modes arranged
uniformly on a ring of radius $4$. All methods start from the same small
Gaussian centered at the origin, away from those modes. For this benchmark we use a
shared multiscale RBF kernel with bandwidth factors $\{0.5,1,2\}$ for all
particle methods so that local spacing and longer-range repulsion are both
represented. This also demonstrates that the multirate formulation is agnostic
to kernel choice: here we use a richer kernel than a single-bandwidth RBF to
capture both local and longer-range interactions. Prior SVGD work has also
considered broader kernel families beyond a single scalar RBF
\cite{wang2019matrixsvgd,stordal2021pkernel}. 

\begin{figure}[t]
  \centering
  \safeincludegraphics[width=0.9\linewidth]{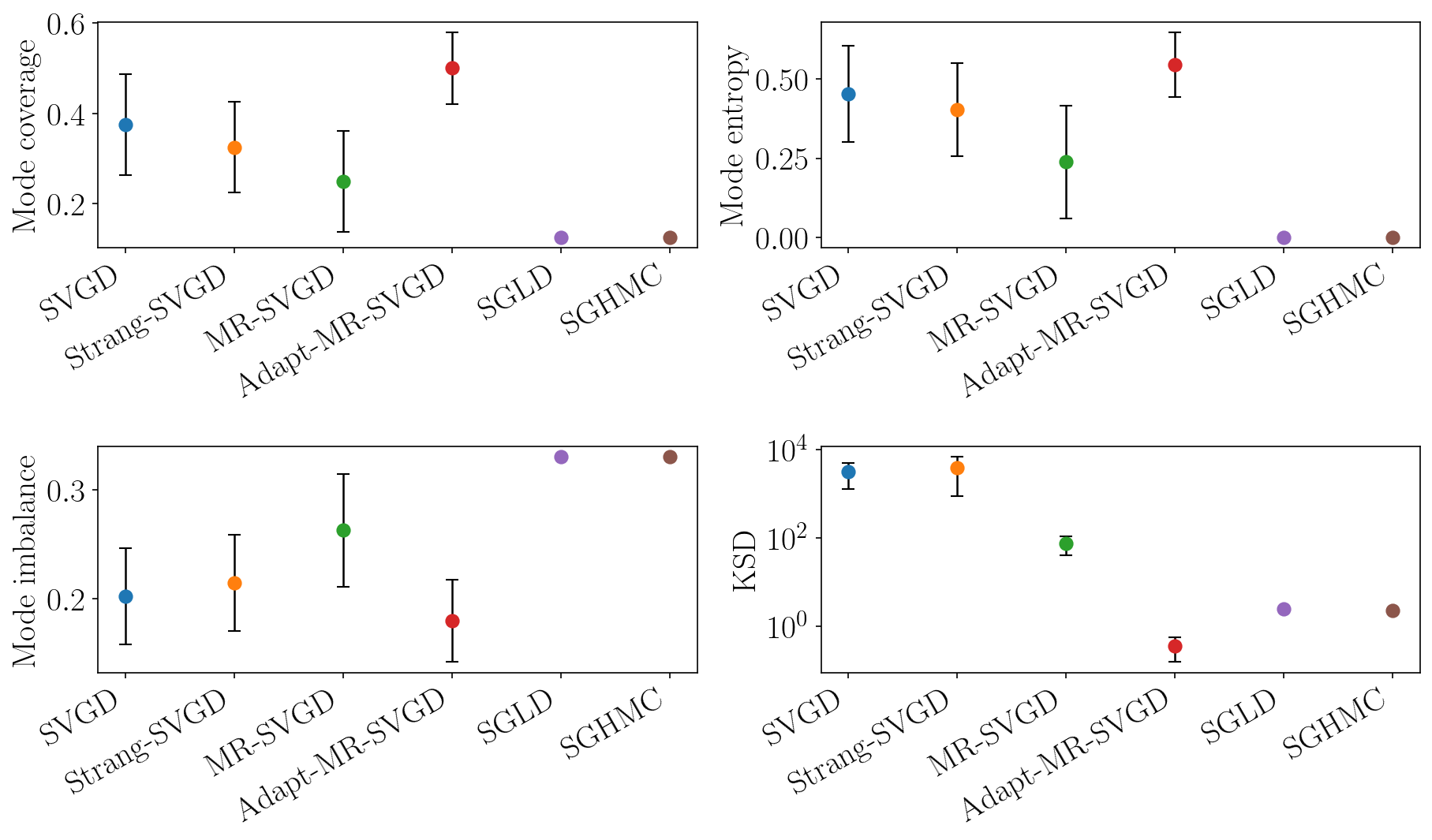}
  \caption{Mixture2D (mix8) fixed-budget final-checkpoint summary across methods.
  The four panels report mode coverage, mode entropy, mode imbalance, and KSD.
  Higher is better for coverage and entropy, while lower is better for mode
  imbalance and KSD.}
  \label{fig:mix8_summary}
\end{figure}

Let $a(x_i)$ denote the nearest
mixture center for particle $x_i$, and let
\[
p_k=\frac{1}{N}\sum_{i=1}^{N}\mathbf{1}\{a(x_i)=k\}
\]
be the empirical mass of mode $k$. We define mode coverage as the fraction of
modes with $p_k\ge 0.05$, normalized mode entropy as
$-\sum_{k=1}^{K} p_k\log p_k / \log K$, and mode imbalance as
$\mathrm{std}(p_1,\ldots,p_K)$. Thus, higher coverage and entropy indicate
broader exploration, while lower imbalance indicates more even mass allocation
across discovered modes. The primary metrics are therefore mode coverage,
normalized mode entropy, and mode imbalance. KSD and mean log probability are
retained as secondary diagnostics. The fixed-budget results are summarized in
\cref{fig:mix8_summary}, which reports the mean final metrics across seeds
together with one standard deviation.

\Cref{fig:mix8_summary} shows that Adapt-MR-SVGD gives the strongest
fixed-budget multimodal result. It attains the highest mean final coverage and
entropy across seeds ($0.500$ and $0.544$, respectively) while keeping KSD at
$0.360$, far below the single-rate particle baselines. Vanilla SVGD and
Strang-SVGD spread farther around the ring than MR-SVGD, but that extra spread
is unstable, with final KSD in the $10^3$ range. Fixed MR-SVGD is much more
stable than those baselines, yet it still underexplores the ring, reaching only
$0.250$ coverage and $0.238$ entropy.

The imbalance panel further shows that Adapt-MR-SVGD distributes mass more
evenly across the modes it discovers. By contrast, SGLD and SGHMC remain near a
single mode, with coverage fixed at $0.125$ and zero entropy, so their small
KSD values do not indicate successful multimodal exploration.

\subsection{UCI logistic regression}

We benchmark Bayesian logistic regression on four UCI datasets
\cite{dua2019uci}: breast cancer, ionosphere, spambase, and a5a. All four are
binary tasks with a Gaussian prior on the regression weights. We emphasize test NLL, accuracy, and ECE, with ESS reported as a
supplemental mixing diagnostic. Summaries are shown in
\cref{fig:uci_summary}, and the particle-method best-NLL summary is reported in
\cref{tab:uci_nll_summary}.  In both we use the best finite-NLL checkpoint selected by
the predictive stopping rule.

\begin{figure}[t]
  \centering
  \begin{subfigure}[t]{0.9\linewidth}
    \centering
    \safeincludegraphics[width=\linewidth]{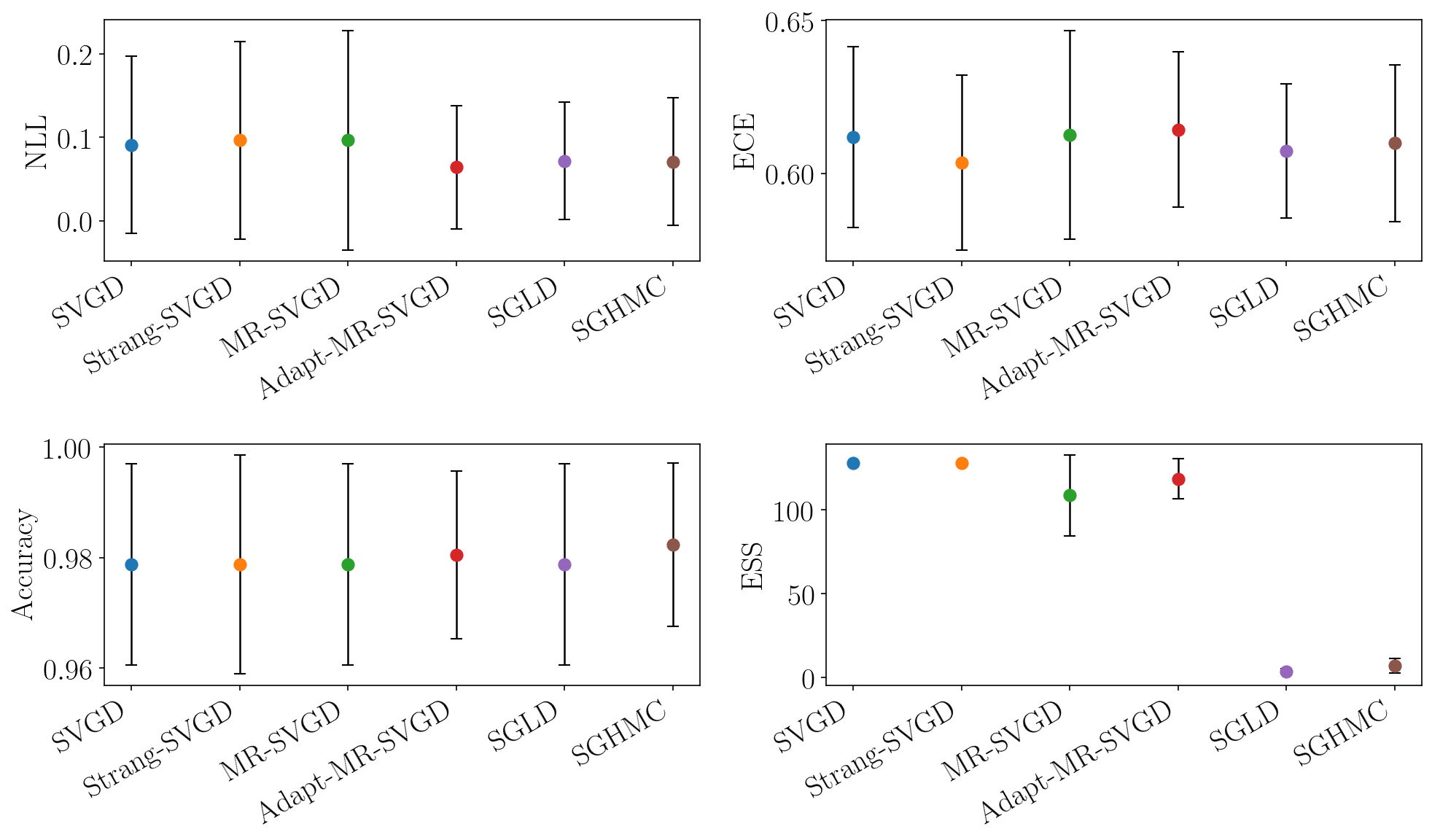}
    \caption{Breast cancer}
    \label{fig:uci_breast_cancer}
  \end{subfigure}

  \vspace{0.25em}

  \begin{subfigure}[t]{0.9\linewidth}
    \centering
    \safeincludegraphics[width=\linewidth]{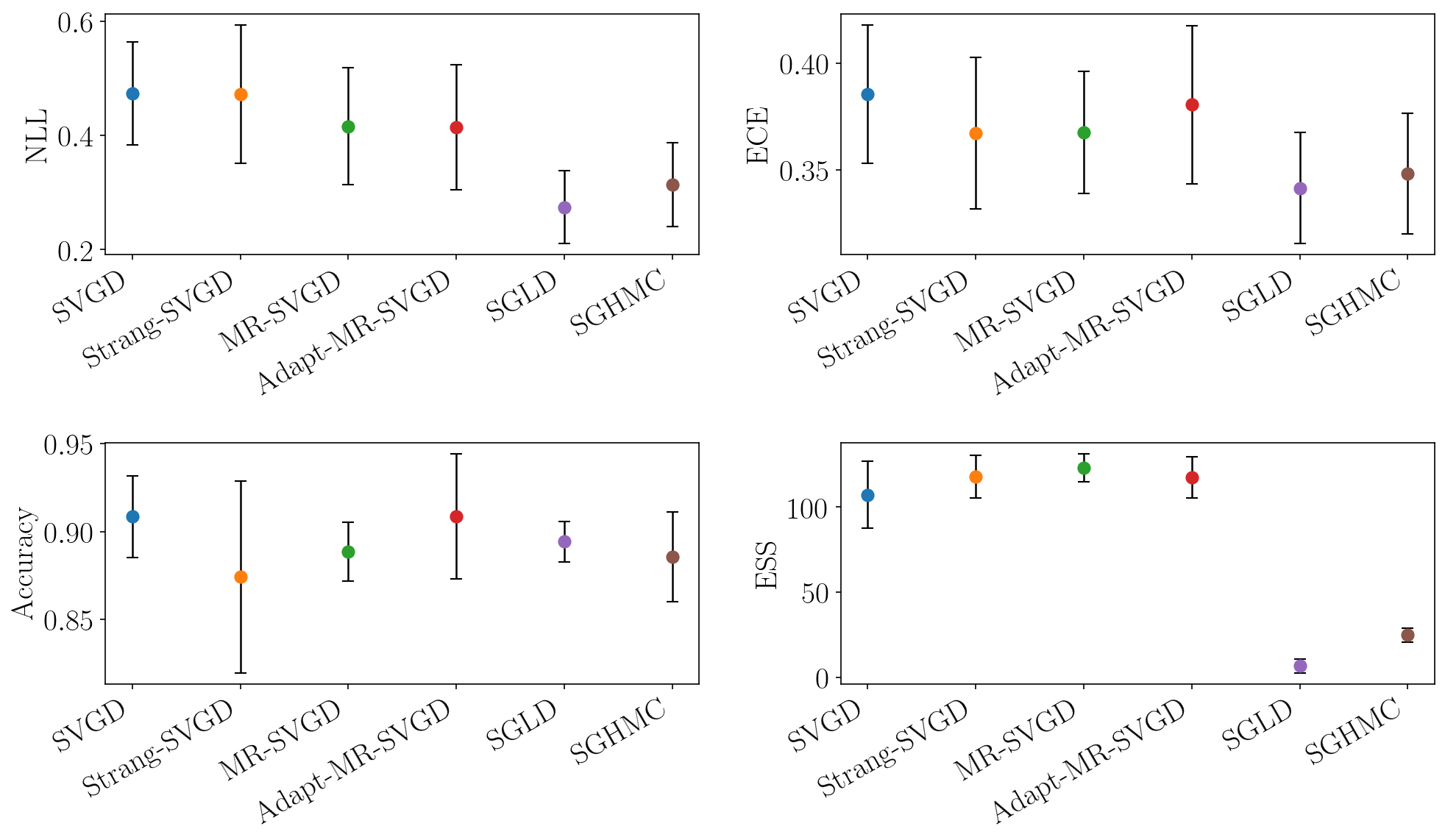}
    \caption{Ionosphere}
    \label{fig:uci_ionosphere}
  \end{subfigure}
  \caption{UCI logistic regression summary across datasets. Each subpanel
  reports test accuracy, NLL, ECE, and ESS at the best finite-NLL checkpoint
  selected by the predictive stopping rule. Remaining datasets are shown in the
  continued panel.}
  \label{fig:uci_summary}
\end{figure}

\begin{figure}[t]
  \ContinuedFloat
  \centering
  \begin{subfigure}[t]{0.9\linewidth}
    \centering
    \safeincludegraphics[width=\linewidth]{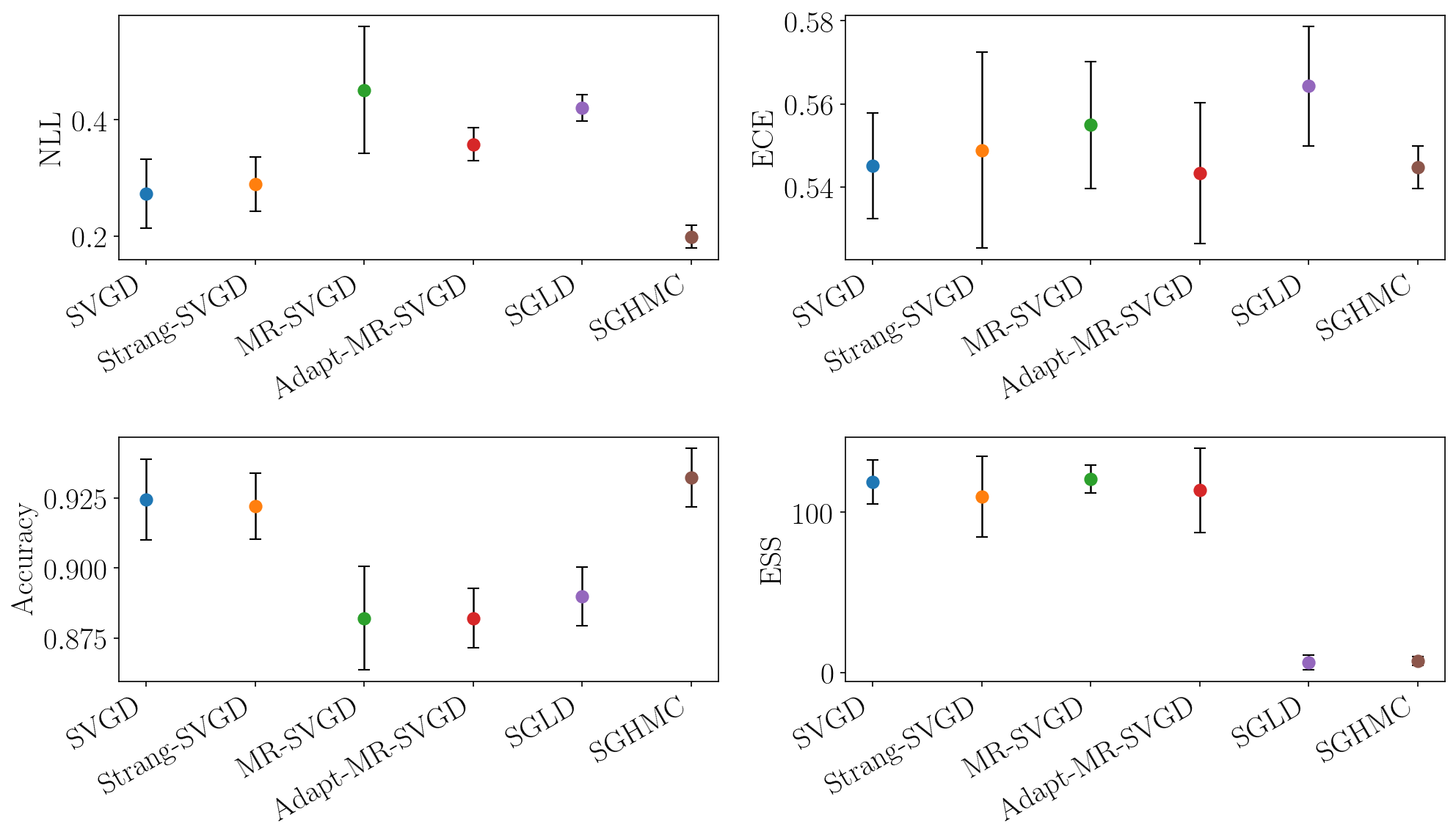}
    \caption{Spambase}
    \label{fig:uci_spambase}
  \end{subfigure}

  \vspace{0.25em}

  \begin{subfigure}[t]{0.9\linewidth}
    \centering
    \safeincludegraphics[width=\linewidth]{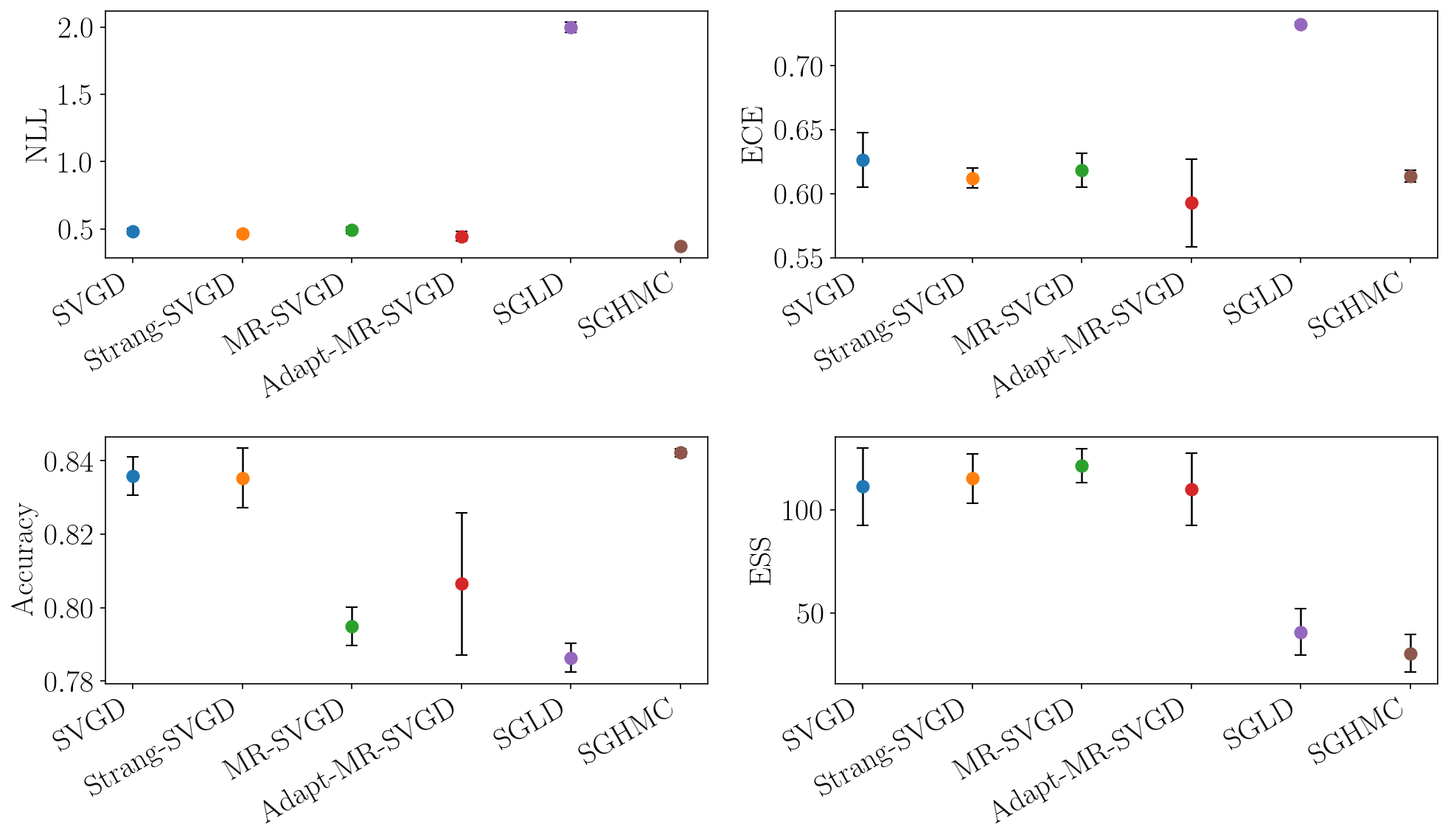}
    \caption{a5a}
    \label{fig:uci_a5a}
  \end{subfigure}
  \caption[]{UCI logistic regression summary across datasets (continued).}
\end{figure}

\begin{table}[t]
  \centering
  \scriptsize
  \setlength{\tabcolsep}{4pt}
  \begin{tabular}{lccc}
    \toprule
    Method & Best NLL$\downarrow$ & Acc@best NLL$\uparrow$ & Wall@best (s)$\downarrow$ \\
    \midrule
    \multicolumn{4}{l}{\textit{Breast cancer}} \\
    Adapt-MR-SVGD & 0.028 & 0.987 & 3.7 \\
    SVGD & 0.091 & 0.979 & 1.4 \\
    Strang-SVGD & 0.096 & 0.979 & 4.0 \\
    MR-SVGD & 0.096 & 0.979 & 3.4 \\
    \midrule
    \multicolumn{4}{l}{\textit{Ionosphere}} \\
    Adapt-MR-SVGD & 0.414 & 0.909 & 5.2 \\
    MR-SVGD & 0.416 & 0.889 & 3.9 \\
    Strang-SVGD & 0.473 & 0.874 & 3.0 \\
    SVGD & 0.473 & 0.909 & 0.6 \\
    \midrule
    \multicolumn{4}{l}{\textit{Spambase}} \\
    SVGD & 0.273 & 0.925 & 0.8 \\
    Strang-SVGD & 0.289 & 0.922 & 3.3 \\
    Adapt-MR-SVGD & 0.358 & 0.882 & 6.6 \\
    MR-SVGD & 0.452 & 0.882 & 3.9 \\
    \midrule
    \multicolumn{4}{l}{\textit{a5a}} \\
    Adapt-MR-SVGD & 0.444 & 0.806 & 14.2 \\
    Strang-SVGD & 0.463 & 0.835 & 2.9 \\
    SVGD & 0.482 & 0.836 & 1.6 \\
    MR-SVGD & 0.500 & 0.793 & 6.6 \\
    \bottomrule
  \end{tabular}
  \caption{UCI logistic-regression particle-method summary. For each dataset
  and method, we average over the finite best-NLL checkpoints selected per seed
  and report NLL, the corresponding accuracy, and wall time. ECE is summarized
  in the per-dataset panels of \cref{fig:uci_summary}.}
  \label{tab:uci_nll_summary}
\end{table}

The particle-only summary in \cref{tab:uci_nll_summary} shows that
Adapt-MR-SVGD attains the best NLL on breast cancer, ionosphere, and a5a,
whereas vanilla SVGD is best on spambase. The broader comparison in
\cref{fig:uci_summary} extends this view to the chain baselines and shows that
SGHMC on spambase and a5a, and SGLD on ionosphere, can achieve lower NLL at
substantially lower wall time, albeit with ESS far below the particle methods.
Accuracy varies little across methods on these datasets, so NLL provides the
more discriminative predictive comparison. Overall, UCI logistic regression is
a comparatively mild benchmark for multirate splitting: adaptive multirate
updates remain competitive, but the clearest multirate gains in this paper
appear on the more anisotropic, multimodal, and hierarchical targets.

\subsection{Bayesian neural network}

\begin{figure}[t]
  \centering
  \begin{subfigure}[t]{0.9\linewidth}
    \centering
    \safeincludegraphics[width=\linewidth]{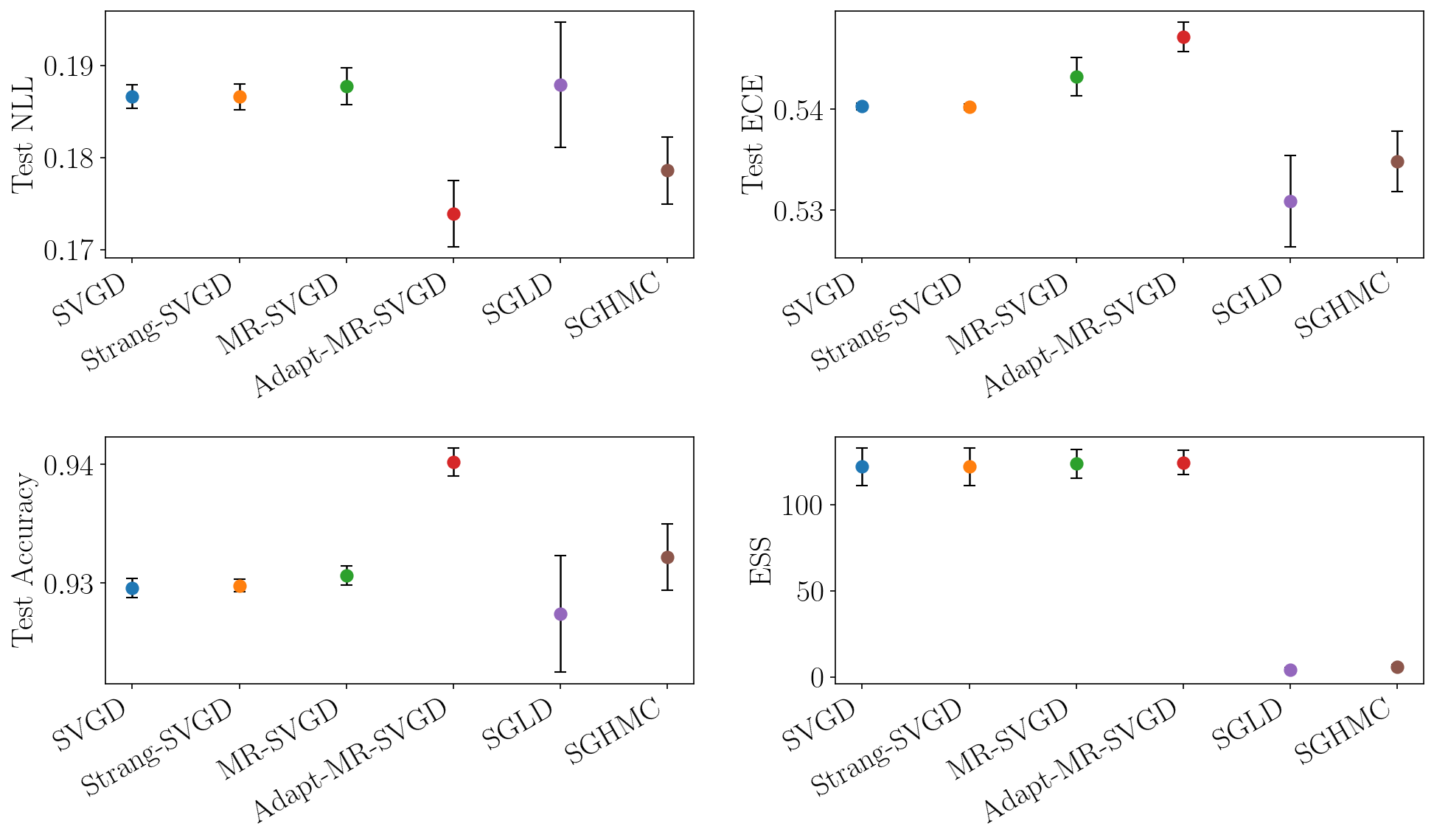}
    \caption{Spambase}
    \label{fig:bnn_spambase}
  \end{subfigure}

  \vspace{0.25em}

  \begin{subfigure}[t]{0.9\linewidth}
    \centering
    \safeincludegraphics[width=\linewidth]{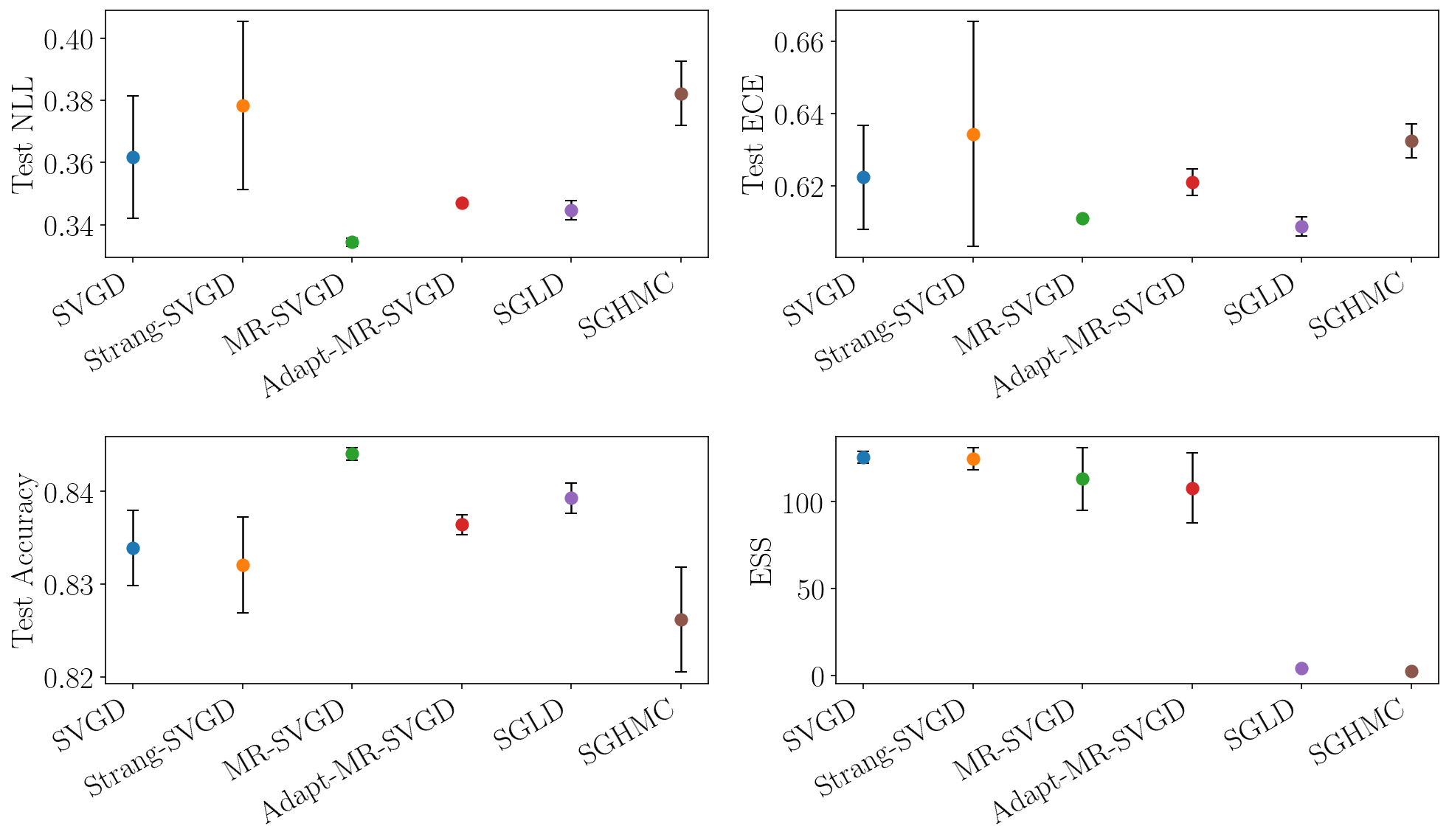}
    \caption{a5a}
    \label{fig:bnn_a5a}
  \end{subfigure}
  \caption{BNN predictive summary across datasets. Each subpanel reports test
  accuracy, NLL, ECE, and ESS at the checkpoint selected by validation NLL.}
  \label{fig:bnn_summary}
\end{figure}

We evaluate a one-hidden-layer Bayesian neural network (width 32) on
Spambase and a5a. We use a Gaussian prior with one shared validation split per
dataset. Checkpoints are selected by validation NLL
under the global predictive stopping rule, and we report test NLL, accuracy,
ECE, ESS, and mean log probability at the selected checkpoint. Per-dataset
summaries appear in \cref{fig:bnn_summary}.

The per-dataset summaries in \cref{fig:bnn_summary} show a split outcome
between the two datasets. On Spambase, Adapt-MR-SVGD attains the best test NLL
and the highest mean test accuracy, while the remaining particle methods
cluster tightly together. On a5a, the ranking shifts: MR-SVGD attains the best
overall test NLL and the highest mean test accuracy, with Adapt-MR-SVGD
remaining the second-best particle method and the single-rate particle variants
again worse. The same panels show that SGHMC on Spambase and SGLD on a5a
remain competitive, but with ESS far below the particle methods.

\subsection{Large-scale hierarchical logistic regression}

To stress-test multirate behavior in a high-dimensional anisotropic posterior,
we study a large-scale hierarchical logistic regression (HLR) benchmark with grouped
random effects. The combination of many group-level effects and a global scale
parameter induces funnel-like geometry and strong local stiffness, making this
benchmark a demanding joint test of predictive quality, robustness, and cost
for the SVGD variants.

\paragraph{Problem definition}
Given binary outcomes $y_i\in\{0,1\}$, sparse features $x_i\in\mathbb{R}^p$, and
group indices $g_i\in\{1,\ldots,G\}$, we consider
\begin{equation}
\begin{aligned}
y_i \mid x_i,g_i,\beta,\alpha,u
&\sim \operatorname{Bernoulli}\!\left(
\sigma\!\left(\alpha + x_i^\top\beta + u_{g_i}\right)\right), \\
&\qquad i=1,\ldots,n,
\end{aligned}
\label{eq:hlr-likelihood}
\end{equation}
where $\sigma(\cdot)$ is the logistic link. We use a non-centered
parameterization for group effects,
\begin{equation}
\begin{aligned}
\beta &\sim \mathcal{N}(0,\sigma_\beta^2 I_p), &
\alpha &\sim \mathcal{N}(0,\sigma_\alpha^2), \\
u_j &= \tau z_j, &
z_j &\sim \mathcal{N}(0,1), \\
\log\tau &\sim \mathcal{N}(\mu_\tau,\sigma_\tau^2),
\end{aligned}
\label{eq:hlr-priors}
\end{equation}
for $j=1,\ldots,G$. The posterior (up to normalization) is
\begin{equation}
\begin{aligned}
\pi(\theta) &\propto
\left[\prod_{i=1}^{n}
\operatorname{Bernoulli}\!\left(
y_i;\sigma\!\left(\alpha+x_i^\top\beta+u_{g_i}\right)\right)\right] \\
&\qquad \times p(\beta)\,p(\alpha)\,p(z)\,p(\log\tau), \\
\theta &= (\beta,\alpha,z,\log\tau).
\end{aligned}
\label{eq:hlr-posterior}
\end{equation}

\paragraph{Protocol}
Our primary HLR benchmark uses a long-tail group-frequency distribution with
$n=10^6$ observations, $p=300$ sparse features, and $G=50{,}000$ groups. Particle methods use 32 particles, whereas single-chain
baselines are evaluated with rolling-window checkpoint summaries. As in the
other predictive benchmarks, checkpoint selection and early stopping are driven
by predictive NLL; here we additionally stop after two consecutive non-finite
checkpoints to prevent unstable runs from drifting far outside the finite-NLL
regime. A uniform-group variant is reported below as a sensitivity check.

\begin{table}[t]
  \centering
  \scriptsize
  \setlength{\tabcolsep}{4pt}
  \begin{tabular}{lcccc}
    \toprule
    Method & Finite seeds$\uparrow$ & Best NLL$\downarrow$ & Acc@best NLL$\uparrow$ & Wall@best (s)$\downarrow$ \\
    \midrule
    Adapt-MR-SVGD & 5/5 & 0.613 & 0.658 & 123.3 \\
    MR-SVGD & 5/5 & 0.623 & 0.656 & 92.5 \\
    SGHMC & 2/5 & 0.860 & 0.560 & 77.9 \\
    SVGD & 3/5 & 0.885 & 0.577 & 154.2 \\
    Strang-SVGD & 3/5 & 1.263 & 0.579 & 163.6 \\
    SGLD & 4/5 & 1.823 & 0.576 & 101.6 \\
    \bottomrule
  \end{tabular}
  \caption{HLR long-tail summary over five seeds. For each seed and method, we
  select the best finite-NLL checkpoint and report method-wise means of that
  NLL, the accuracy at the selected checkpoint, and the corresponding wall
  time. ``Finite seeds'' counts runs that reached at least one finite-NLL
  checkpoint before stopping. This is the main hierarchical stress test:
  Adapt-MR-SVGD attains the lowest mean best-NLL while remaining finite on all
  seeds, and MR-SVGD is the closest lower-cost baseline.}
  \label{tab:hlr_longtail_summary}
\end{table}

\paragraph{Result summary}
\cref{tab:hlr_longtail_summary} shows the clearest predictive separation among
the particle methods in the paper. Adapt-MR-SVGD attains the lowest mean
best-NLL ($0.613$) while remaining finite on all five seeds. MR-SVGD is the
closest competitor, with only slightly worse NLL ($0.623$) and a lower wall
time at the selected checkpoint (92.5\,s versus 123.3\,s). By contrast,
vanilla SVGD, Strang-SVGD, and SGHMC frequently fail to reach stable finite
checkpoints in this regime, while SGLD remains finite more often but with
substantially worse predictive performance. In the long-tail setting, the main
distinction is therefore not only final NLL, but whether the method remains
numerically viable under the hierarchical funnel geometry.

\begin{table}[t]
  \centering
  \scriptsize
  \setlength{\tabcolsep}{4pt}
  \begin{tabular}{lcccc}
    \toprule
    Method & Finite seeds$\uparrow$ & Best NLL$\downarrow$ & Acc@best NLL$\uparrow$ & Wall@best (s)$\downarrow$ \\
    \midrule
    Adapt-MR-SVGD & 5/5 & 0.681 & 0.601 & 94.6 \\
    SVGD & 1/5 & 0.700 & 0.548 & 54.9 \\
    MR-SVGD & 5/5 & 0.706 & 0.597 & 44.2 \\
    Strang-SVGD & 1/5 & 0.708 & 0.584 & 241.7 \\
    SGLD & 5/5 & 0.743 & 0.528 & 66.0 \\
    SGHMC & 5/5 & 0.791 & 0.554 & 79.7 \\
    \bottomrule
  \end{tabular}
  \caption{HLR uniform-group summary over five seeds. Entries are defined as in
  \cref{tab:hlr_longtail_summary}, but with balanced group assignments as a
  sensitivity check.}
  \label{tab:hlr_uniform_summary}
\end{table}

\paragraph{Uniform-group sensitivity check}
\cref{tab:hlr_uniform_summary} confirms the same qualitative ordering under a
less pathological grouping structure. Adapt-MR-SVGD again attains the lowest
mean best-NLL with full finite-seed coverage, while MR-SVGD remains close in
NLL and is the fastest among methods with full finite coverage. The long-tail
regime remains the more discriminative benchmark, but the relative ranking of
the multirate variants is stable across both group-assignment patterns.

\section{Conclusions}
\label{sec:conclusions}

This work studies a central numerical weakness of SVGD: a single global step
size must simultaneously resolve attractive drift and diversity-preserving
repulsion even when these components evolve on different effective time scales.
We address this mismatch by recasting SVGD as a split interacting-particle flow
and integrating the drift and repulsion terms with multirate updates.

Across a broad benchmark suite, the main empirical pattern is that multirate
SVGD improves robustness and quality-cost tradeoffs relative to vanilla SVGD,
with the strongest gains on anisotropic, multimodal, and hierarchical targets.
Fixed multirate schedules provide a simpler robustness baseline, while adaptive
drift resolution offers the strongest performance when local stiffness changes
along the trajectory. At the same time, no single variant dominates every
metric and every regime, reinforcing the importance of evaluating quality and
cost jointly rather than through a single summary statistic.

Several limitations remain. Current adaptivity is centered on drift resolution,
while repulsion frequency is still controlled by simple schedules. ESS remains a
univariate proxy, and dense kernel interactions continue to constrain scaling at
large particle counts. These observations suggest clear next steps: tighter
adaptive controllers that coordinate drift and repulsion updates, approximate or
structured kernel computations for large-$N$ regimes, and stronger theoretical
links between multirate stability analysis and particle-based Bayesian
inference guarantees. Together, these directions can move multirate SVGD from a
robust empirical strategy toward a more principled and scalable foundation for
modern Bayesian computation.

\section*{Data availability}

Code, experiment scripts, and manuscript sources for this study are maintained
in a public GitHub repository.\footnote{\url{https://github.com/csml-beach/multirate-sampling}}
Public benchmark inputs used in the predictive experiments are derived from
standard UCI datasets. All other benchmark targets are synthetic and are
generated directly by the repository code. Generated figures, metrics, and
supplementary animation artifacts are available in the project repository for
reference.

\section*{Declaration of generative AI and AI-assisted technologies}

The \texttt{JAX} codebase for this project was developed in part with the assistance of OpenAI Codex. After using this tool, the author reviewed and edited the content as needed and takes full responsibility for the content of the published article.
\ifblindsubmission
\else
\section*{Acknowledgments}

This research was supported by the CSML lab and the College of Engineering at
California State University, Long Beach. The work used Jetstream2 at Indiana
University through allocation CIS230277 from the Advanced
Cyberinfrastructure Coordination Ecosystem: Services \& Support (ACCESS)
program, which is supported by National Science Foundation grants \#2138259,
\#2138286, \#2138307, \#2137603, and \#2138296.

\fi

\bibliographystyle{elsarticle-num}
\bibliography{Bib/references,Bib/sarshar,Bib/sandu,Bib/ode_multirate,Bib/ode_imex,Bib/ode_general}

\end{document}